\documentclass{ifacconf}

\usepackage{graphicx}      
\usepackage{natbib}        
\usepackage{amsmath,amssymb,amsfonts}
\usepackage{algorithm}
\usepackage{algpseudocode}
\usepackage{graphicx}
\usepackage{xcolor}
\usepackage{subcaption}
\usepackage{bm}
\usepackage[acronym,nomain,nonumberlist]{glossaries}

\newacronym{uav}{UAV}{Unmanned Aerial Vehicle}
\newacronym{sar}{SAR}{Search and Rescue}
\newacronym{flie}{FLIE}{First Look Inspect-Explore}
\newacronym{bim}{BIM}{Building Information Model}
\newacronym{cad}{CAD}{Computer Aided Design}
\newacronym{soa}{SoA}{State-of-Art}
\newacronym{nn}{NN}{Nearest Neighbour}
\newacronym{mav}{MAV}{Micro Aerial Vehicle}

\begin{document}
\begin{frontmatter}

\title{Towards a Reduced Dependency Framework for Autonomous Unified Inspect-Explore Missions}


\author[First]{Vignesh Kottayam Viswanathan},
\author[First]{Sumeet Gajanan Satpute},
\author[Second]{Ali-akbar Agha-mohammadi} and
\author[First]{George Nikolakopoulos}

\address[First]{Robotics and AI Team\\ Lule\aa\,\,University of Technology, Lule\aa\, Sweden \\ (e-mail: vigkot@ltu.se, sumsat@ltu.se and geonik@ltu.se)}
\address[Second]{AI for Humanity Inc, (e-mail: ali.agha4@gmail.com)}

\thanks[footnoteinfo]{This work has been partially funded by the European Union’s Horizon 2020 Research and
Innovation Programme under the Grant Agreement No. 101003591 NEXGEN SIMS}

\begin{abstract}                

The task of establishing and maintaining situational awareness in an unknown
environment is a critical step to fulfil in a mission related to the field of rescue robotics. Predominantly, the problem of visual inspection of urban structures is dealt with view-planning being addressed by map-based approaches. In this article, we propose a novel approach towards effective use of Micro Aerial Vehicles (MAVs) for obtaining a 3-D shape of an unknown structure of objects utilizing a map-independent planning framework. The problem is undertaken via a bifurcated approach to address the task of executing a closer inspection of detected structures with a wider exploration strategy to identify and locate nearby structures, while being equipped with limited sensing capability. The proposed framework is evaluated experimentally in a controlled indoor environment in presence of a mock-up environment validating the efficacy of the proposed inspect-explore policy.
\end{abstract}


\end{frontmatter}

\section{Introduction}
The field of autonomous aerial robots has seen significantly challenging deployment scenarios in recent times. With rising potential of on-board autonomy integrated with limited sensing hardware on the \glspl{mav}, the use of such low-cost, highly-effective platform is aimed to serve the research niche of maximizing information gain through reduced hardware capability while serving to fulfil the objectives of missions such as \gls{sar} operations ~\cite{lindqvist2022compra,alotaibi2019lsar}, rapid navigation of underground systems~\cite{mansouri2020subterranean,mansouri2018towards} and executing infrastructure inspections~\cite{quenzel2019autonomous,eudes2018autonomous}. Thus, these platforms are intrinsic towards minimizing the overall cost of deployment and off-setting the danger to human life in high-risk situations, while ensuring that the desired mission objectives are met.

\begin{figure}[htpb]
    \centering
    \includegraphics[width = \linewidth]{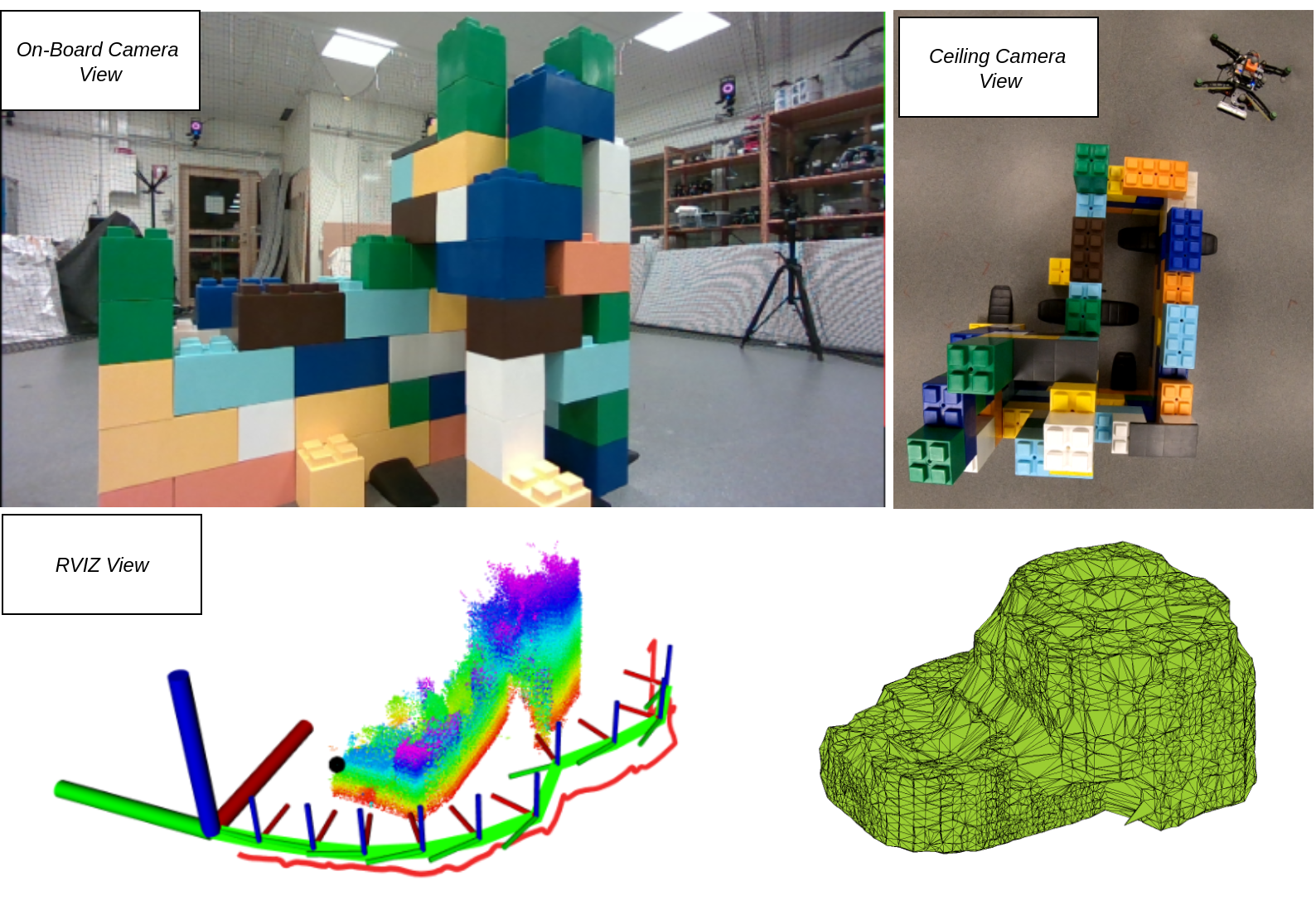}
    \caption{(\textit{On Top Left}) A snapshot of the video from the on-board camera taken during the inspect-explore mission is shown. (\textit{On Top Right}) A captured shot of the \gls{mav} from a GoPRo mounted on the ceiling above the structure. (\textit{On Bottom Left}) A snapshot of RVIZ view of the \gls{mav} taken during the inspect-explore run. (\textit{On Bottom Right}) A post-processed 3D alphashape generated from the merged point-cloud detected during the experiment.}
    \label{fig:concpet_fig}
\end{figure}

Enabling autonomous inspection of a-priori known complex infrastructures, through the use of aerial agents, has for long been a coveted application, as it is a preferred substitute to complete labour-intensive tasks, mainly due to the associated advantages, such as ease-of-deployment, task repeatability and accurate data collection. Moreover, in recent times, the applications envelope include challenging situations, such as inspection of previously unknown structures with the \gls{mav} operating in an unknown or contaminated environment~\cite{song2017online,naazare2022online}, thus, contributing towards improving the situational awareness of the local region. Figure~\ref{fig:concpet_fig} presents the proposed FLIE autonomy in action.

The proposed work addresses key components falling under the perspective of rescue robotics, focused specifically towards visual inspection of fractured and discontinuous structures in a GPS-denied region using a resource-constrained platform. The inspiration behind the proposed \gls{flie} framework is to realize a flexible behaviour tree that ensures a safe and autonomous \gls{mav} operation in an unknown environment. In the method presented, the task of improving situational awareness is bifurcated into dual branches: 1. A primary inspection node, which aims to provides a closer and detailed information of the structure through recursive view planning based on the First-Look methodology~\cite{viswanathan2022first}, subjected to photogrammetric constraints and, 2. A secondary exploration node, which locates the presence of nearby structures through a multi-tiered hierarchical search policy.

\subsection{Background \& Motivation}

Towards the inspection of unknown structures ~\cite{naazare2022online} presented a modified form of the Next-Best-View method,~\cite{connolly1985determination}. In their work, the authors implement a volumetric map-based heuristic approach through the consideration of weighted-sum for the information gain obtained to drive the decision to explore or inspect to address specific needs in the field of disaster response. From the work of~\cite{brogaard2021towards}, the problem of inspection-exploration in a constrained environment, such as the water ballast tanks of ships, is addressed. Through an octomap built using point-cloud from the stereo camera, the authors propose Rapidly-exploring Random Tree (RRT) methodology for implementing collision-free navigation during exploration and closer defect identification and observation via an on-board neural network.~\cite{choi2021online} address the task of coverage planning of bridge based on an occupancy-map constructed from depth point-cloud. The authors presented a combination of Principal Component Analysis (PCA) and k-Nearest Neighbour (k-NN) search to provide view vector orientations to map the bridge surface.  

The works of~\cite{song2017online} and ~\cite{song2020online} employed a \gls{mav} to construct 3D models of previously unknown environments based on Frontier driven Next Best View planner. In~\cite{song2020online}, the authors proposed a sector decomposition strategy on a spatially bounded map to provide informative view poses to cover local frontiers within a given sector, while ~\cite{song2017online} proposed a sampling-based optimization problem to satisfy th desired coverage, to be achieved by planning an optimal path, through a sequence of planned view configurations.~\cite{bircher2018receding} presented a sampling-based Receding Horizon NBV (RH-NBV) that ensures optimal planning of sensor configurations for exploration of an unknown volume and inspection of a desired surface based on an occupancy-map.

\subsection{Contributions}
The proposed FLIE autonomy package focuses on improving the situational awareness of a previously unknown environment, through a novel implementation of a unified inspect-explore framework. This work addresses the dual aspect of providing information-rich spatial map of an unknown environment through 3D reconstruction, i.e by performing close inspection of structures surveyed to be present, via a tiered exploration strategy of the environment in the vicinity of the UAV. The main contributions of the work are as follows:
\begin{enumerate}
    \item  We introduce a novel vision-based unified inspect-explore autonomy framework, coupled with multi-tiered hierarchical exploration strategy that addresses the need to identify and navigate towards nearby structures located around the vicinity of the UAV.
    \item A novel feature based scene recognition module is implemented during the inspection task, in addition to a profile adaptive view-planner, which dynamically maintains the required inspection distance through a framework integrated passive collision-avoidance scheme.
    \item We perform an experimental evaluation of the proposed autonomy to prove the efficacy of the overall framework.
\end{enumerate}

The article is structured as follows. Section~\ref{Sec1} defines the problem addressed in this work along with explanation of preliminaries. In Section~\ref{Sec2}, the proposed methodology is presented. Section~\ref{Sec3} describes, in brief, the low level autonomy implemented in this work. The experimental and \gls{mav} setup carried out for this work is given in Section~\ref{Sec4}. Section~\ref{Sec5} presents the results obtained including a comprehensive discussion of the experimental outcome. Finally, Section~\ref{Sec6} concludes the article through a brief summary of the presented work and the findings obtained in addition to offering a few directions of future work that can be carried out.
\section{Modelling and Problem Statement} \label{Sec1}

We consider an unbounded world space $\bm{\mathcal{V}} \in \mathbb{R}^3$ filled with fractured urban structures, represented in the form of a collection of points $\bm{\mathcal{S}} \in \mathbb{R}^3$, detected by the on-board optical sensing hardware. The objective of the proposed framework is to survey, inspect and generate a bounded 3D shape of available structures in the global space $\bm{\mathcal{M}} \in \mathbb{R}^3| \bm{\mathcal{M} \supseteq \bm{\mathcal{S}}}$ by enabling the UAV to navigate the unbounded obstructed landscape, through the stored repository, containing visited poses $\bm{\mathcal{\xi}} \in \mathcal{R}^4$ generated based on the \textit{First-Look} approach in a recursive manner, until there exists no new structures within the vicinity of the explored world space.

\subsection{Frames of reference}
\begin{figure}[htpb]
    \centering
    \includegraphics[width = 0.8\linewidth]{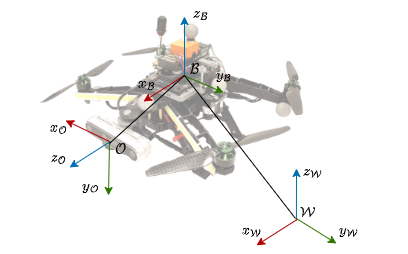}
    \caption{Graphical representation of applicable reference frames utilized in the presented work}
    \label{fig:FoR}
\end{figure}

Let $\bm{\mathcal{W} \in \mathbb{R}^3}$ be the fixed global frame of reference with the UAV pose, given as
$\bm{\xi}_{mav} = [\Vec{p}_{mav},~\psi_{mav}]$, where $\Vec{p}_{mav} = [x,~y,~z] \in \mathbb{R}^3$. Let $\bm{\mathcal{B}}\in\mathbb{R}^3$ the body-fixed frame of the UAV to which a
stereo camera is attached with the optical frame given as $\bm{\mathcal{O}}\in\mathbb{R}^3$.
Figure~\ref{fig:FoR} graphically depicts the layout of the applicable reference in the presented work.

\section{FLIE Autonomy} \label{Sec2}

\begin{figure*}[htpb]
    \centering
    \includegraphics[width = 0.8\linewidth]{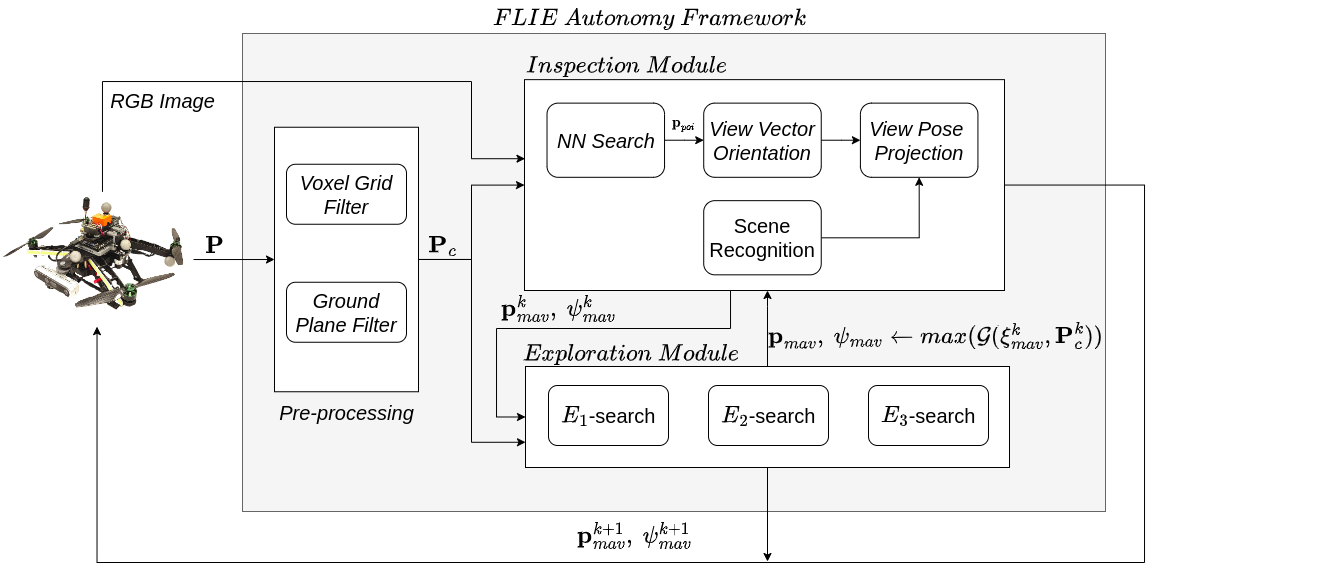}
    \caption{Graphical representation of the proposed FLIE autonomy framework.}
    \label{fig:FLIE_frmwk}
\end{figure*}

 Figure~\ref{fig:FLIE_frmwk} depicts a schematic representation of the proposed FLIE autonomy. In this Section, we present the bifurcated architecture consisting of inspection and exploration modules implemented to enable a \gls{mav} to perform visual inspection of an unknown structure. While the inspection framework provides profile adaptive view poses, subject to photogrammetric constraints and maintaining a safe viewing distance, the exploration module addresses the need to locate and navigate towards remaining structures that are present in the vicinity of the inspected structure through a tiered hierarchical search policies. Furthermore, the proposed framework depends solely on the optical sensor on-board to carry-out view planning. 

\subsection{Inspection Framework} \label{insp_mod}


Leveraging the inspection planner in ~\cite{viswanathan2022first}, the
inspection framework implemented here considers the depth point-cloud information
$\Vec{P}_c \in \mathbb{R}^3$ from the on-board stereo camera. Subsequent to which, a 
k-dimensional tree is generated to find the nearest point $\Vec{p}_{poi} \in \mathbb{R}^3$
with respective to the \gls{mav}'s current position. 

In addition to dynamically re-configuring the view poses based on the current sensor information, inspecting fractured or discontinuous structures often requires a low-level collision-avoidance scheme to account for obstructive projections from the structure. To this extent, a dual-purpose safety-layer is integrated into the baseline framework, which considers the result of nearest-neighbour search, executed on $\bm{P}_c$, thereby directly influencing the required view pose. Thus, simultaneously maintaining the desired inspection distance from the structure, while actively securing the safety of the \gls{mav}, as it travels around the structure and accounting for any path-obscuring protrusions from the structure that fall within the field-of-view of the optical sensor.


Equation.~\eqref{fl_insp_eqn} represents the mathematical formulation of the aforementioned safety-layer integrated with the view planner, where $k \in \mathbb{R}$ is the progressive iteration, being executed inside the framework. As it can be inferred, both the view planner and the collision-avoidance scheme primarily depend on the current normalized view vectors of the \gls{mav} along $\mathbf{X}, \mathbf{Y}$ and $\mathbf{Z}$ directions given by $\Vec{v}_x , \Vec{v}_y, \Vec{v}_z \in \mathbb{R}^3$. 

\begin{equation} \label{fl_insp_eqn}
    \Vec{p}^{k+1|k}_{mav} =  \Vec{p}^{k|k}_{mav} + \Vec{v}_{y,k} \mathcal{O}_v + \Vec{v}_{x,k} \Vec{d}_{safety}
\end{equation}

\begin{equation}\label{yaw_ref}
    \psi^{k+1|k}_{mav} = \textit{arctan}(\Vec{v}_x(2),\Vec{v}_x(1))
\end{equation}
where,
\begin{equation*}
    \Vec{v}_x = \frac{\Vec{p}_{poi}-\Vec{p}^{k}_{mav}}{||\Vec{p}_{poi}-\Vec{p}^{k}_{mav}||}
\end{equation*}
\begin{equation*}
    \Vec{v}_y = \textit{cross}(\Vec{n}_z,\Vec{v}_x), \Vec{n}_z \in \mathbb{R}^3
\end{equation*}
\begin{equation*}
    \Vec{v}_z = \textit{cross}(\Vec{v}_y,\Vec{v}_x)
\end{equation*}

Therefore, in a recursive manner, the updated view pose satisfying horizontal overlap $\mathcal{O}_h \in \mathbb{R}$ condition is projected along the current $\Vec{v}_y$ direction, while a safe inspection distance is maintained along the current $\Vec{v}_x$ direction. This formulation enables the \gls{mav} to configure it's inspection behaviour based on the locally viewed surface profile of the structure. Equation.~\eqref{horz_ovl} and \eqref{vert_ovl} provide the required horizontal and vertical overlapping distance based on the current position of the \gls{mav}, the nearest detected neighbour, the desired overlap factor $\gamma_h,\gamma_v \in \mathbb{R}$ and the field of view of the camera $\alpha$,$\beta \in \mathbb{R}$ along its horizontal and vertical axis respectively.

\begin{equation}\label{horz_ovl}
\mathcal{O}_h = f(\Vec{p}^{k}_{mav}, \Vec{p}_{poi}, \alpha, \gamma_h)
\end{equation}
where, 
\begin{align*}
    f(\Vec{p}^{k}_{mav}, p_{poi}, \alpha, \gamma_h) &= 2\textit{tan}\frac{\alpha}{2}||\Vec{p}_{poi} - \Vec{p}^k_{mav} ||\\& -2\gamma_h\textit{tan}\frac{\alpha}{2}||\Vec{p}_{poi} - \Vec{p}^k_{mav}||
\end{align*}
Similarly, the required vertical overlap can be written as,
\begin{equation}\label{vert_ovl}
\mathcal{O}_v = f(\Vec{p}^{k}_{mav}, \Vec{p}_{poi}, \beta, \gamma_v)
\end{equation}
where,
\begin{align*}
    f(\Vec{p}^{k}_{mav}, \Vec{p}_{poi}, \beta, \gamma_v) &= 2\textit{tan}\frac{\beta}{2}||\Vec{p}_{poi} - \Vec{p}^k_{mav} ||\\& -2\gamma_v\textit{tan}\frac{\beta}{2}||\Vec{p}_{poi} - \Vec{p}^k_{mav}||
\end{align*}
In addition to considering horizontal overlap constraints between two successive inspection view points, the planner also takes into account the required vertical overlap $\mathcal{O}_v \in \mathbb{R}$, which is projected along the $\Vec{v}_z$ direction. Thus, the inspection planner follows a sequential format to provide a visual coverage of an unknown structure, that is, the planner executes a horizontal inspection loop around the structure, while addressing the required view orientation with respect to the local surface being viewed. Following which, it commands the \gls{mav} to move onto the next vertical level, based on the required $\mathcal{O}_v$ and continues the process in a recursive manner, until it inspects the entire structure. Algorithm.~\ref{alg:insp_mod} presents the pseudo-code of the implemented inspection planner.
 
\begin{algorithm}
	\caption{Visual Inspection}
	\label{alg:insp_mod}
	\begin{algorithmic}[1]
	\While {\textbf{not} inspected}
	    \If {$\emptyset \xleftarrow{}$ \textbf{P}}
	        \State Proceed to Explore
	    \Else
    	    \State $\mathbf{P}_c\xleftarrow{}$ \textbf{VoxelGridFilter}($\mathbf{P}$)  
    	    \State $\textit{visited\_scene}$ $\xleftarrow{}$ \textbf{SceneRecog}($\mathcal{I}_q$)
    	    \State $\vec{p}_{poi}\xleftarrow{}$ \textbf{NNsearch}($\mathbf{P}_c$~,~$\Vec{p}^k_{mav}$)
    	    \State $[\Vec{v}_x~\Vec{v}_y~\Vec{v}_z] \xleftarrow{}$ \textbf{ViewOrientation}($\Vec{p}_{poi},\Vec{p}^k_{mav}$)
    	    \If {$visited\_scene$}
    	        \State $\Vec{p}^{k+1}_{mav} =\Vec{p}^k_{mav}~+~\Vec{v}_y\mathcal{O}_{h}~+~\Vec{v}_{x,k} \Vec{d}_{safety}~+~\Vec{v}_z\mathcal{O}_{v}  $
    	    \Else
    	        \State $\Vec{p}^{k+1}_{mav} =\Vec{p}^k_{mav}~+~\Vec{v}_y\mathcal{O}_{h}~+~\Vec{v}_{x,k} \Vec{d}_{safety} $
    	    \EndIf
    	    \If {argmax(\textbf{P}$_c$[:~,~3]) $<$ ~$\Vec{p}^{k+1}_{mav}[3]$}
    	        \State Return to Base and Explore 
    	   \Else 
    	        \State \textbf{Continue}
    	    \EndIf
    \EndIf
	\EndWhile
	\end{algorithmic} 
\end{algorithm}%
\subsubsection{Scene Recognition} 

The primary objective of this module is to provide a quantitative measure of understanding the scene containing the current structure under inspection. By considering the RGB image from the on-board stereo camera, the scene recognition module assists the inspection planner to execute the condition of required vertical overlap by identifying the associated image frames from the previously inspected surfaces of the structure. As \gls{mav} proceeds to inspect the target structure, Oriented Binary Robust Independent Elementary Features (ORB) descriptors~\cite{rublee2011orb} are used to identify unique matches $\mathcal{F}\in\mathbb{R}$ between the candidate image $\mathcal{I}_c$ and the query image $\mathcal{I}_q$ captured by the \gls{mav} at the beginning of the inspection loop. Let $\gamma_{similarity} \in \mathbb{R}$ be the calculated similarity score obtained from the number of matches ($n_{matches}\in\mathbb{Z}^+$), obtained between $\mathcal{I}_c$ and $\mathcal{I}_q$ and the number of original descriptors ($n_{descriptors}\in\mathbb{Z}^+$), derived from $\mathcal{I}_c$. Equation~\eqref{eqn:sim_score} refers to the mathematical model implemented to address scene recognition with $N$ corresponding to the desired horizon to set the mean similarity threshold. To account for deployment in various environments, the threshold value for $\gamma_{similarity}$ is not pre-defined, instead, an average score is calculated from the arithmetic mean over a horizon of a user-defined number of image frames and is dynamically set as the threshold score to identify scenes of previously inspected surfaces, irrespective of the operating environment. 
\begin{equation}\label{eqn:sim_score}
    \gamma_{similarity} =  \frac{n_{matches}}{n_{descriptors}} \geq \Bar{\gamma}_{threshold}
\end{equation}
where,
\begin{equation*}
    \Bar{\gamma}_{threshold} = \frac{1}{N}\sum^{N}_{i=1}\gamma_{similarity}, N\in\mathbb{Z}^+
\end{equation*}
\subsection{Exploration Framework}

Inspection of a-priori unknown structures often requires a precursor assumption in the form of bounded spatial region of operation or presence of a single solitary structure to carry out the visual coverage. In the proposed work, we forego any such presumptions of the operating environment and focus on maximizing the information gain $\mathcal{G}$(\textbf{P}$_c$) through a tiered search policy. This module aims to direct the \gls{mav} towards the direction of maximum detected $\mathbf{P}_c$ to continue the visual inspection. There are three main search policies modeled: \textbf{E}$_1$ , \textbf{E}$_2$ and \textbf{E}$_3$, each addressing the objective to find nearby structures based on conditions, such as previous engagement of inspection of a structure, available $\textbf{P}_c$ information and previously visited view points. 

In the \textbf{E}$_1$ policy, the \gls{mav} is commanded to cover view directions that lie within it's forward view space subject to the stereo camera's field-of-view. \textbf{E}$_1$ policy is executed when the \gls{mav} has shown immediately prior inspection behaviour. This policy aims to address surface discontinuities on structures, such as broken-down walls to enable efficient re-engagement of inspection behaviour preventing the mission from terminating early. Thus, for \textbf{E}$_1$($\Vec{p}^k_{mav}$,$\psi^k_{mav}$) Eqn.~\eqref{E1_expl} presents the commanded directions to investigate potential continuation of surface from current position. 

\begin{equation} \label{E1_expl}
    \psi^{k+1|k}_{mav} = \psi^k_{mav} + \alpha, ~\forall~k \in [1...m] 
\end{equation}
where, $\alpha \in \mathbb{R}$ is the horizontal field-of-view of the on-board camera, $m \in \mathbb{R}$ is the number of search index equal to $\frac{\pi}{0.5 \alpha}$ rounded off to the nearest integer. In the event $\phi \xleftarrow[]{}\mathcal{G}$, i.e no surfaces are seen within the forward view space, the \textbf{E$_2$} search policy provides $360^o$ view coverage from the current pose of the \gls{mav} accounting for any structures present in its immediate vicinity. The modeled \textbf{E$_2$}($\Vec{p}^k_{mav}$,$\psi^k_{mav}$) policy follows similar formulation as presented in ~\eqref{E1_expl} except with modifications to $m$, where during \textbf{E$_2$}-phase, $m \in \frac{2\pi}{\alpha}$ rounded off to the nearest integer. 

If $\phi \xleftarrow[]{}\mathcal{G}$ at the end of \textbf{E$_2$} phase, a travelling search is executed by the exploration module. The \textbf{E$_3$} module aims to address the need to actively search for potential presence of structures located in the region around the current object under inspection through backtracking. \textbf{E$_3$}-phase is initialized when the inspection planner finds \textbf{P}$_c$[:~,~3] $< ~\Vec{p}^{k+1}_{mav}[3]$, i.e  when there exists no detected point-cloud points exceeding the altitude of the next commanded view pose. This would suggest that the \gls{mav} has reached the final feasible inspection view-point around the structure leading to the termination of the inspection planner and commanding the \gls{mav} to return to it's base position. Let $\xi_{insp} = [\Vec{p}_{mav},\psi_{mav}] \in \mathbb{R}^4$ be an ordered sequence of previously planned view poses. Equation.~\eqref{E3_expl_1} represents the backtracking formulation implemented in this work. In~\eqref{E3_expl_1}, $j$ corresponds to the length of the commanded inspection pose at the base level of the inspection loops.  %
\begin{equation}\label{E3_expl_1}
    \xi^{k+1|k}_{mav} = \xi^{j-k|k}_{mav},j \in \mathbb{Z}^+
\end{equation}
where, the view orientation is directed to be $180^o$ offset as shown below: 
\begin{equation}\label{E3_expl_2}
    \psi^{k+1|k}_{mav} =  \pi + \psi^{j-k|k}_{mav}
\end{equation}

The reference heading angles obtained from~\eqref{E1_expl} and~\eqref{E3_expl_2} are bounded within $[-\pi~\pi]$. Algorithm~\ref{alg:exp_mod} presents the pseudo-code of the implemented exploration module.

\begin{algorithm}[htbp]
	\caption{Exploration}
	\label{alg:exp_mod}
	\begin{algorithmic}[1]
    \While{explore}

        \If{engaged} \hfill/* Check for prior engagement of inspection behaviour */
        
            \State Execute \textbf{E}$_1$ policy
            \State \textbf{return} $\xi^{E1}_{mav} \in \max(\mathcal{G})$~\textit{if}~ \textbf{not}~$\phi~\xleftarrow~\mathcal{G}$
        
        \Else \hfill /* Proceed to E$_2$ */
        \State Execute \textbf{E$_2$} policy
        
        \State \textbf{return} $\xi^{E_2}_{mav}$ $\in$ max($\mathcal{G}$)~\textit{if}~\textbf{not} $\phi~\xleftarrow~\mathcal{G}$~\textbf{else}
        \State Proceed to \textbf{E$_3$}
        
        \State \textbf{return} $\xi^{E_3}_{mav}$ $\in$ max($\mathcal{G}$)~\textit{if}~\textbf{not} $\phi~\xleftarrow~\mathcal{G}$~\textbf{else}
        \State \textbf{break} \hfill/* Terminate mission */
        
    \EndIf
    \EndWhile 
	\end{algorithmic} 
\end{algorithm}%

\section{Low Level Autonomy} \label{Sec3}

\begin{figure}[htpb]
    \centering
    \includegraphics[width = 0.8\linewidth]{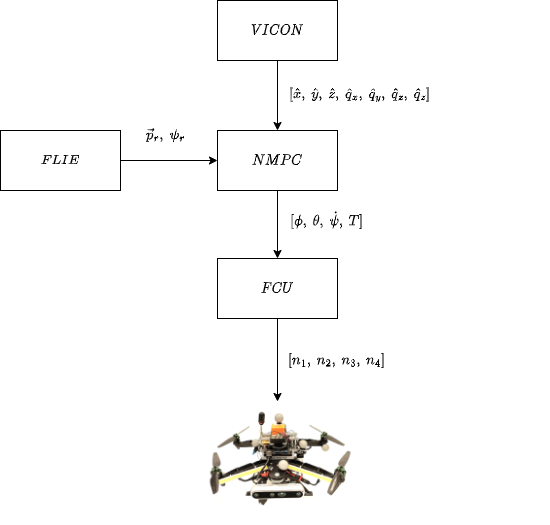}
    \caption{The novel control framework implemented during the experimental trials.}
    \label{fig:ctrl_frmwk}
\end{figure}

In addition to the proposed framework, the \gls{mav} is equipped with a Nonlinear Model Predictive Controller (NMPC) to provide control inputs in the form of angle and thrust commands $u = [\phi,~\theta~,\Dot{\psi},T]$ based on the position $\vec{\overrightarrow{p}}_r = [x,~y,~z]$ and the orientation $\psi_r$ references provided by the FLIE module to the Flight Control Unit (FCU) present on-board. A comprehensive discussion on the implemented baseline controller can be found in~\cite{lindqvist2022adaptive}. The FCU then translates the control references to individual motor speed commands $[n_1,~n_2~n_3~n_4]$ for the \gls{mav}. The indoor positioning service is provided by the \textit{Vicon} motion capture system (MoCAP), which provides estimated state vector $\Vec{\hat{x}} = [\hat{x},~\hat{y},~\hat{z},~\hat{q}_x,~\hat{q}_y,~\hat{q}_z,~\hat{q}_z]$ of the \gls{mav} through tracking of the reflective spheres placed on its body. Figure.~\ref{fig:ctrl_frmwk} represents the controller architecture implemented in this work. 

\section{Experimental Set-up} \label{Sec4}

\begin{figure}[htpb]
    \centering
    \includegraphics[width = \linewidth]{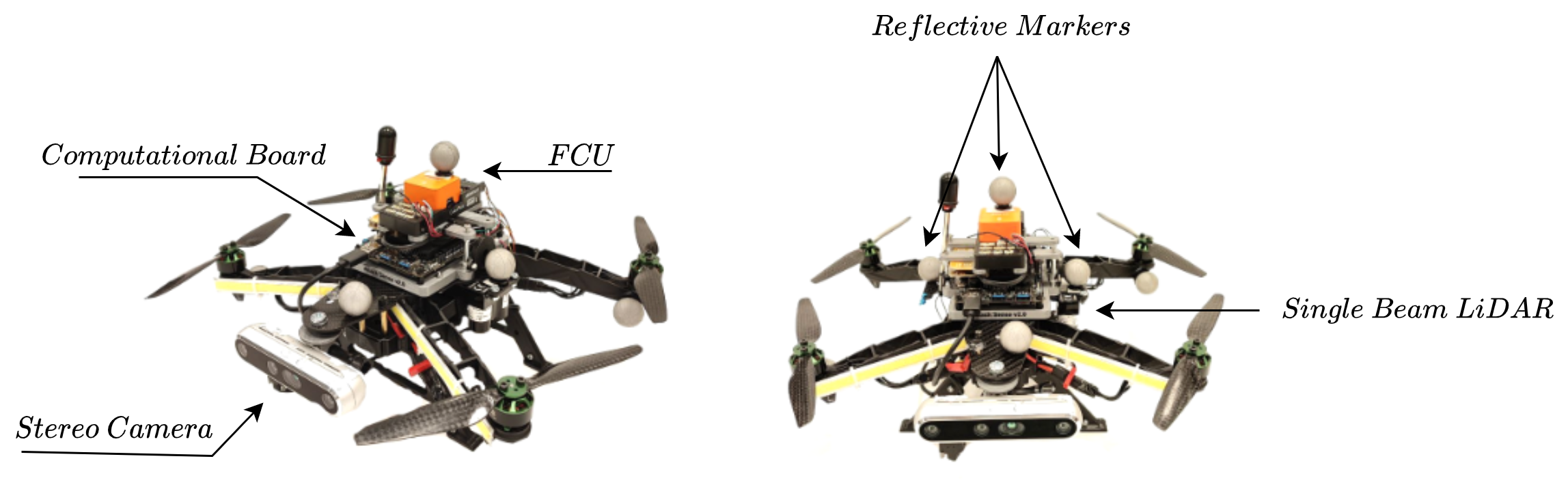}
    \caption{Pictographical depiction of the low-cost \gls{mav} platform along with the on-board sensor-suite deployed in this work.}
    \label{fig:pixy_exp}
\end{figure}

\begin{figure}[htpb]
    \centering
    \includegraphics[width = \linewidth]{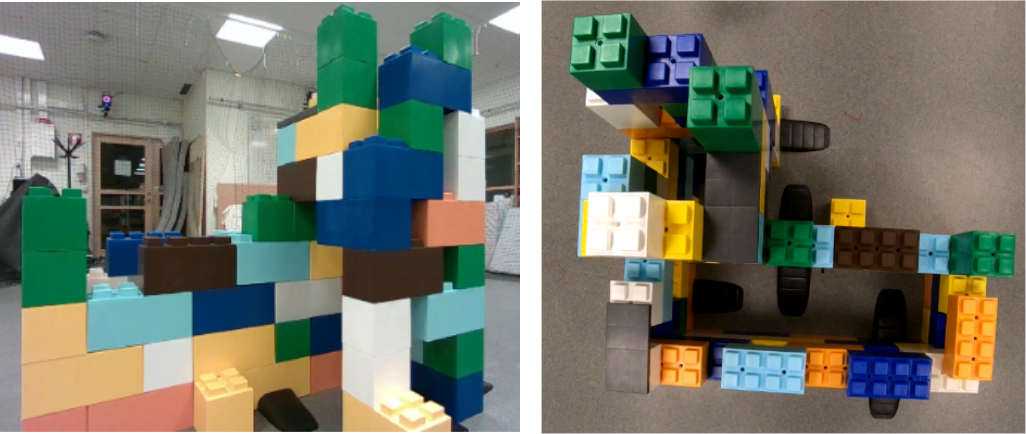}
    \caption{Constructed mock-up of a discontinuous structure using \textit{EverBlock} modular blocks in an indoor environment. }
    \label{fig:env_setup}
\end{figure}

Figure~\ref{fig:pixy_exp} presents the \gls{mav} hardware utilized in this work. The platform's development is presented in~\cite{mansouri2020deploying}. The \gls{mav} is equipped with an on-board \textit{Realsense} $D455$ sensor operating at 30 Hz to provide depth point-cloud and RGB image information, a downwards-facing single beam Lidar-lite v3 operating at 100hz to measure range to the ground. \textit{PixHawk} is implemented as the Flight Control Unit (FCU) on the \gls{mav} with \textit{Lattepanda} Gen3 computational board to process data. Figure~\ref{fig:env_setup} depicts the mock-up structure utilized for the experimental evaluation.

\begin{table}[htpb]
\caption{The initialization parameters implemented for FLIE.}
\centering
\begin{tabular}{|c|l|l|l|l|l|}
\hline
$\alpha$ & $\beta$                      & $\gamma_h$               & $\gamma_v$               & d$_{safety}$             & $\gamma_{threshold}$     \\ \hline
86 $\deg$   & \multicolumn{1}{c|}{57 $\deg$} & \multicolumn{1}{c|}{0.8} & \multicolumn{1}{c|}{0.5} & \multicolumn{1}{c|}{1 m} & \multicolumn{1}{c|}{0.6} \\ \hline
\end{tabular}
\label{tab:experimental parameters}
\end{table}

Table~\ref{tab:experimental parameters} shows the initialization parameters implemented for the experimental evaluation. Since the experiment is conducted indoors, utilization of a voxel grid filter enables the FLIE autonomy to focus on foreground objects and removes $Inf$ and $NaN$ points from the stereo depth cloud. Through multiple experimental trials, the voxel grid has been tuned to maximum sensor range of 1.5$m$ and a nearest visible distance to 0.5$m$ along the viewing direction to address the presence of noise in the depth cloud output from the camera for smaller ranges. In addition to that, the horizontal view space of the camera is limited between -1$m$ and 1$m$. To remove ground plane from the depth cloud, a vertical mask of 0.2$m$ was implemented during run-time.

Due to constraints regarding limited operational space for the \gls{mav} in the indoor environment, the \gls{mav} is made to execute the FLIE mission around a single target structure. For initialization, the mission starts with the \gls{mav} oriented towards the empty space. A video of the experimental evaluation can be found here https://youtu.be/PMb1rsC69rc .
\section{Results and Discussion} \label{Sec5}

\begin{figure}[htpb]
    \centering
    \includegraphics[width = \linewidth]{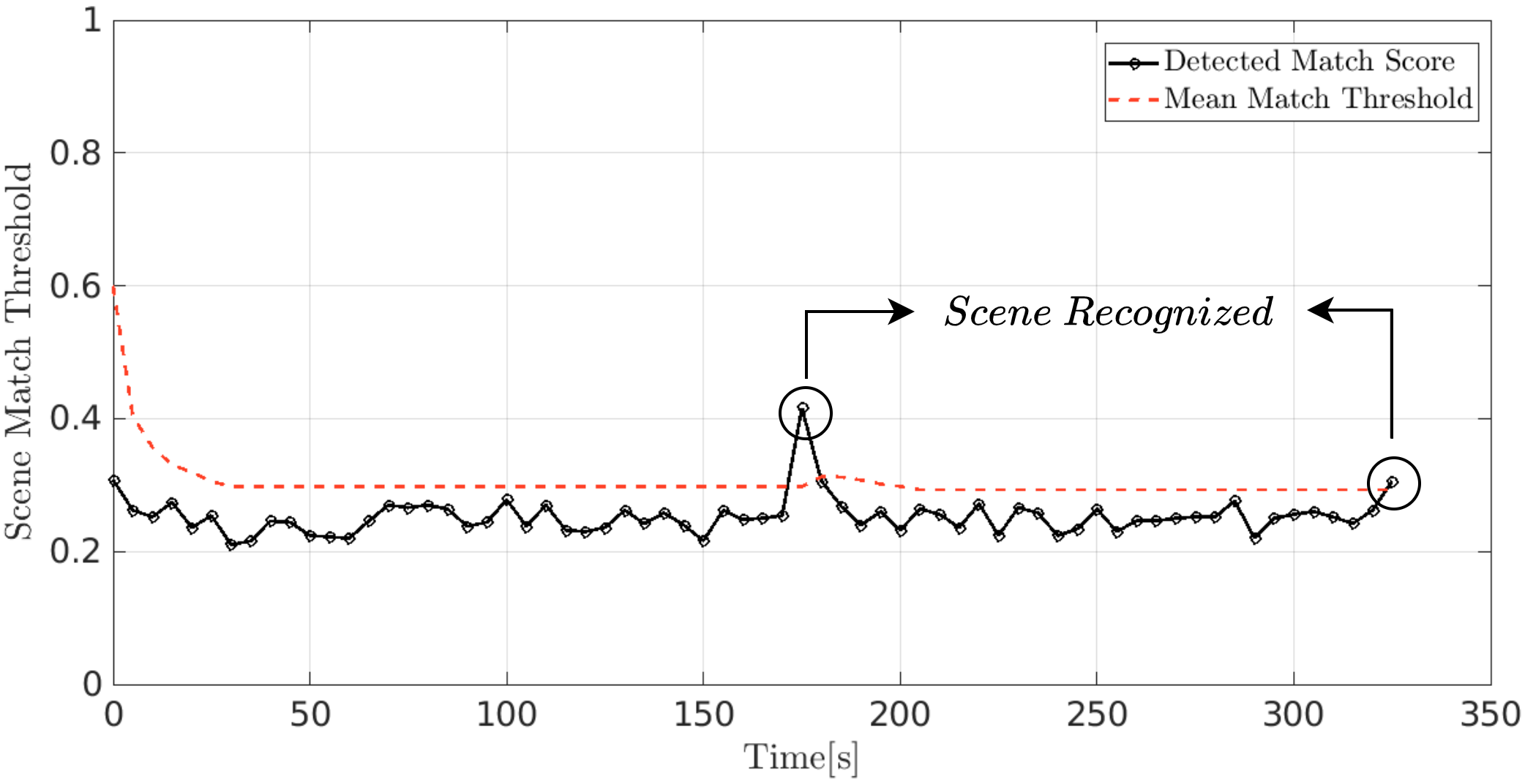}
    \caption{Graphical plot depicting the performance of the scene recognition module during visual inspection of the EverBlock structure. The data-points demarcated indicate the cross-over of feature-matching score with the dynamically set threshold factor. }
    \label{fig:fig_sceneRecog}
\end{figure}

Figure.~\ref{fig:fig_sceneRecog} demonstrates the performance of the scene recognition module implemented in this work to aid in visual inspection. As modelled, the initial value of $\bar{\gamma}_{threshold}$ progressively reduces from $0.6$ and settles at $0.310$ after a horizon of $N=6$. The highlighted data-points on the plot signifies the execution of the required vertical overlap by the planner, as it recognizes previously inspected surfaces after completing an inspection loop. Subsequent to the demarcated point at the right extreme of the plot, inspection is terminated as the \gls{mav} has reached the last feasible point of inspection and proceeds to explore based on \textbf{E$_3$} policy.

\begin{figure}[htpb]
    \centering
    \includegraphics[width = \linewidth]{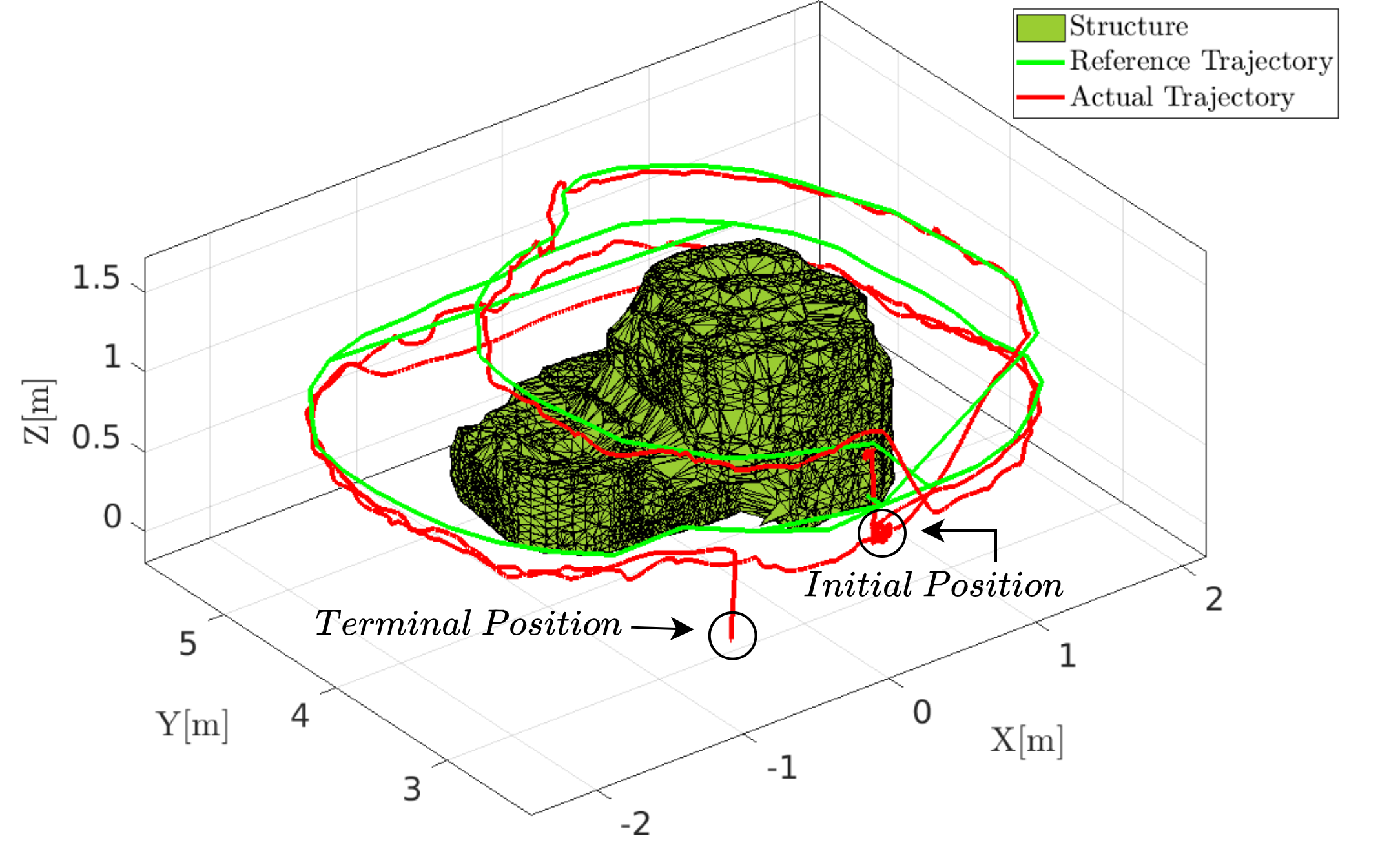}
    \caption{3D visualization of the commanded inspection trajectory (\textit{green}) overlayed with the actual odometry (\textit{red}) tracked by the \gls{mav} during the experimental trials.}
    \label{fig:3d_plot}
\end{figure}

In Fig.~\ref{fig:3d_plot}, a 3D visualization of the executed inspection trajectory around the structure in an indoor environment is provided. The \gls{mav} can be seen following the desired inspection view poses. In addition to that, the implemented FLIE autonomy can be seen dynamically adapting the required inspection path based on the profile of the structure seen. Along with Fig.~\ref{fig:fig_sceneRecog}, the \gls{mav} can be seen to be commanded to reach the base position at the end of the final inspection loop due to absence of any detected structures above its commanded position.
\begin{figure}[htpb]
    \centering
    \includegraphics[width = \linewidth]{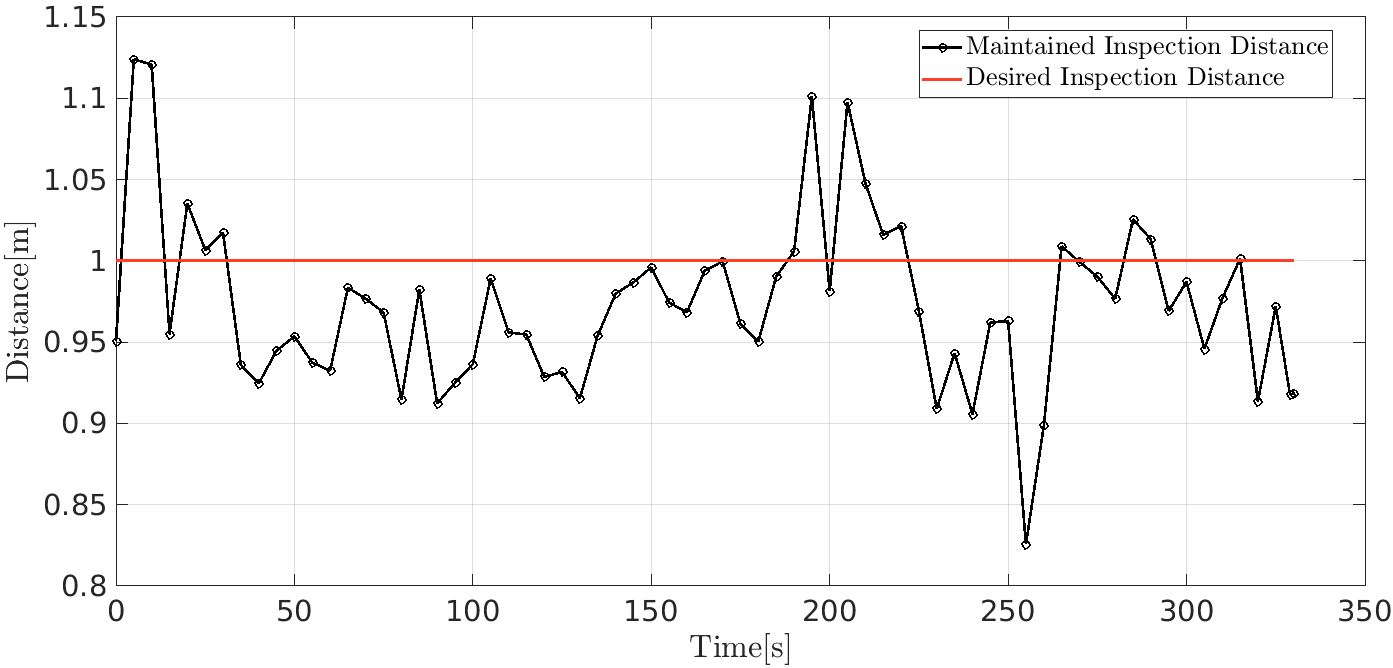}
    \caption{The performance of the dual-purpose safety layer implemented in the inspection planner is displayed with the maintained viewing distance (\textit{black}) against the desired viewing distance (\textit{red}).}
    \label{fig:inspDist}
\end{figure}

\begin{figure}[htpb]
    \centering
    \includegraphics[width = \linewidth]{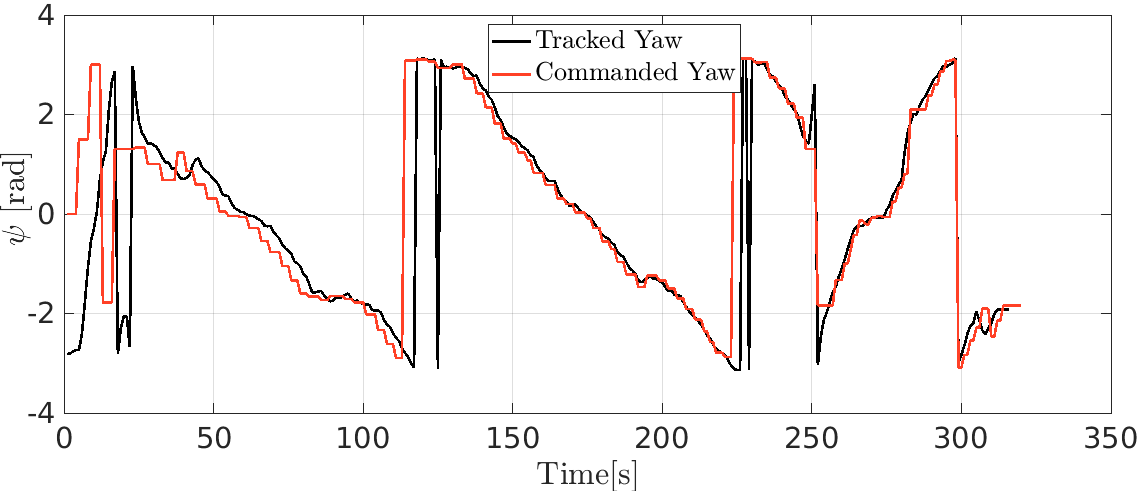}
    \caption{Graphical plot of the commanded heading (\textit{red}) against actual heading (\textit{black}) tracked by the \gls{mav} during the experiment.}
    \label{fig:flie_yaw}
\end{figure}

Figure.~\ref{fig:inspDist} captures the behaviour of the implemented dual-purpose safety layer during the inspection run. Compared to the desired inspection distance of 1$m$, the \gls{mav} can be seen exhibiting the desired response with a mean error of 0.0464$m$. Figure.~\ref{fig:flie_yaw} shows the heading reference commanded by the inspect-explore framework, shown in \textit{red} against the tracked heading by the \gls{mav}, shown in \textit{black}.

\begin{figure}[htpb]
    \centering
    \includegraphics[width = \linewidth]{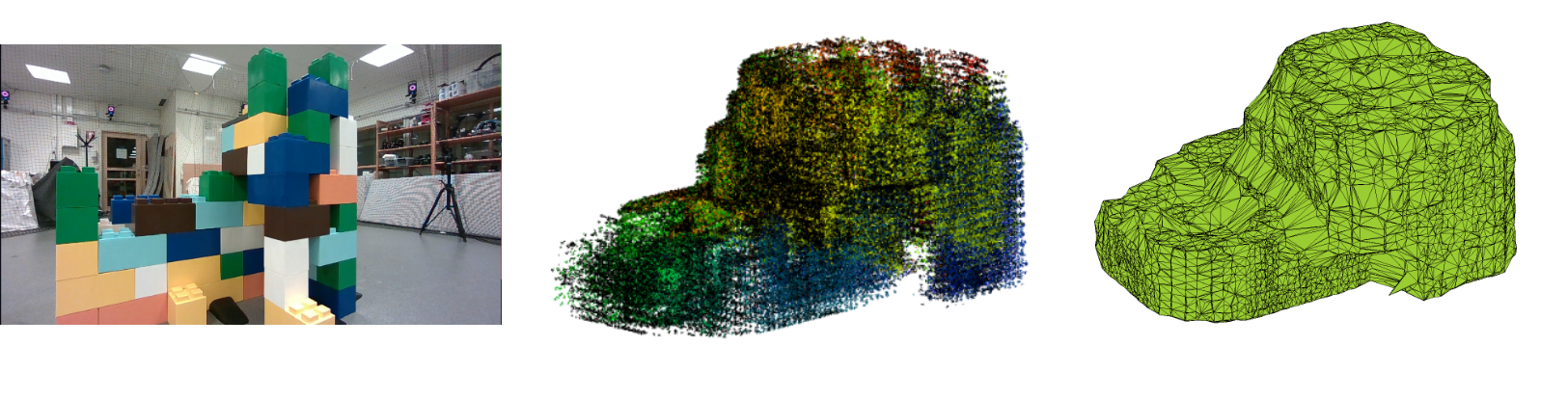}
    \caption{(\textit{On the left}) The RGB image captured from the on-board camera is shown. The constructed EverBlock structure can be seen within the view frame. (\textit{On Center}) Post-processed point-cloud using \textit{CLOUDCOMPARE} obtained during the run is seen. (\textit{On the right}) The 3D alphashape produced from the processed point-cloud information is displayed using \textit{MATLAB}.}
    \label{fig:flie_pp}
\end{figure}


Figure~\ref{fig:flie_pp} shows a snapshot from the on-board camera taken during the experiment (left-side), along with the final merged depth cloud captured (center), and the 3D shape generated from the merged cloud (right). \textit{CLOUDCOMPARE} was used to process the merged cloud and \textit{MATLAB} was used to generate the final alpha-shape representation.
An inspected volume of 2.1326 $m^3$ is obtained from the 3D object. Due to small errors during transformations, from \textbf{P}$^\mathcal{B}_c$ to \textbf{P}$^\mathcal{W}_c$, through the indoor positioning system utilized, it can be seen that the final obtained merged point-cloud has drifted from the original structure representation. Compared to map-based approaches for inspect-explore missions, despite the accumulated drift in the 3D map, the proposed FLIE autonomy is resilient to such occurrences due to the planner operating in an recursive fashion directly based on the current available sensor information. Thus, avoiding view planning on progressively built map.

\section{Conclusions} \label{Sec6}
In this work, a novel First-Look based Inspect-Explore Autonomy for improving situational awareness of a structure of unknown objects is presented. The proposed framework is implemented using a unified approach towards inspection and exploration to enable the \gls{mav} to progressively survey and closely inspect structures located in its vicinity. In addition to that, the framework integrates a feature-based scene recognition module to address errors in positioning system, enabling the \gls{mav} to execute its experiment successfully by recognizing previously inspected surfaces, thus being resilient to any drift that would arise. The framework is also verified to be independent of the need of a map to execute view-planning. Experimental evaluations performed in an indoor environment demonstrates the efficacy of the developed inspect-explore method.

\bibliography{ifac}
\end{document}